\documentclass[conference]{IEEEtran}
\IEEEoverridecommandlockouts
\usepackage[numbers]{natbib}
\usepackage{amsmath,amssymb,amsfonts}
\usepackage{algorithmic}
\usepackage{graphicx}
\usepackage{textcomp}
\usepackage{xcolor}
\usepackage{array}
\usepackage{url}
\usepackage{arydshln}
\usepackage{subcaption}
\usepackage{multirow}
\usepackage{hyperref}
\usepackage{caption}%c
\def\BibTeX{{\rm B\kern-.05em{\sc i\kern-.025em b}\kern-.08em
    T\kern-.1667em\lower.7ex\hbox{E}\kern-.125emX}}
    
\begin{document}

\title{Enhancing Stance Classification on Social Media Using Quantified Moral Foundations}

\author{
    \IEEEauthorblockN{Hong Zhang$^{1, *}$, Quoc-Nam Nguyen$^{1, *}$, Prasanta Bhattacharya$^2$, Wei Gao$^1$, Liang Ze Wong$^2$, Brandon Siyuan Loh$^2$, \\Joseph J. P. Simons$^2$, Jisun An$^3$}
    \IEEEauthorblockA{
        \textit{$^1$School of Computing and Information Systems, Singapore Management University, Singapore}\\
        hong.zhang.2022@phdcs.smu.edu.sg, \{qnnguyen, weigao\}@smu.edu.sg
    }
    \IEEEauthorblockA{
        \textit{$^2$Institute of High Performance Computing (IHPC), Agency for Science, Technology and Research (A*STAR)}\\
        \textit{1 Fusionopolis Way, \#16-16 Connexis, Singapore 138632, Republic of Singapore}\\
        \{prasanta\_bhattacharya, wong\_liang\_ze, brandon\_loh, simonsj\}@ihpc.a-star.edu.sg
    }
    \IEEEauthorblockA{
        \textit{$^3$Luddy School of Informatics, Computing, and Engineering, Indiana University Bloomington, IN, USA}\\
        jisun.an@acm.org
    }
    \thanks{$^*$These authors contributed equally to this work.}
}

\maketitle

\begin{abstract}
This study enhances stance detection on social media by incorporating deeper psychological attributes, specifically individuals' moral foundations. These theoretically-derived dimensions aim to provide a comprehensive profile of an individual's moral concerns which, in recent work, has been linked to behaviour in a range of domains, including society, politics, health, and the environment. In this paper, we investigate how moral foundation dimensions can contribute to predicting an individual's stance on a given target. Specifically we incorporate moral foundation features extracted from text, along with message semantic features, to classify stances at both message- and user-levels using both traditional machine learning models and large language models. Our preliminary results suggest that encoding moral foundations can enhance the performance of stance detection tasks and help illuminate the associations between specific moral foundations and online stances on target topics. The results highlight the importance of considering deeper psychological attributes in stance analysis and underscores the role of moral foundations in guiding online social behavior.
\end{abstract}

\begin{IEEEkeywords}
stance detection, morality foundations, user behaviour
\end{IEEEkeywords}

\section{Introduction}
%\textcolor{red}{Prasanta: Let's first start with an opening para talking about (a) what is stance, (b) stance detection as a task, (c) gaps in current stance detection work, and then transition to a discussion on moral foundations, as below.}\\
As a behaviour, a \textit{stance} refers broadly to an expression of perspectives, attitudes, or judgments toward a given proposition. The study of stances is inherently interdisciplinary and traces its roots to psycho- and socio-linguistics. In their popular text, \citet{biber1988adverbial} define stance as the ``lexical and grammatical expression of attitudes, feelings, judgments, or commitment concerning the propositional content of a message.''. Similarly, \citet{du2007stance} refers to stance as ``an articulated form of social action'', which involves evaluation of an object, or an alignment with a given position.

As a communicative media, social media offers a perfect context for the expression and study of user stances at scale. Recent studies have proposed models for detecting or inferring the stance conveyed in social media posts on target topics \cite{aldayel2021stance, kuccuk2020stance}. While the vast majority of these studies have leveraged message-level characteristics such as language use and user interactions, the question of whether stance modeling can be improved through the incorporation of richer user-level attributes, notably psychological characteristics, remains understudied, with only a few recent exceptions \cite{rezapour2021incorporating}.  

%\textcolor{red}{Prasanta: This para should introduce briefly the evolution of moral foundations as a theory, and then mainly discuss \textbf{why} incorporating moral foundations should be useful for detecting stances}

Our working hypothesis is that a user's broader value system carries relevant information for inferring their stance on a particular topic. 
%If one wants to know what a person thinks on a given issue, it is useful to know not only what they say about that issue but also their broader moral commitments and concerns. 
In this study, we operationalize users' values using Moral Foundations Theory \cite{haidt2004intuitive,graham2011mapping}. Drawing on cross-cultural and evolutionary psychological research \cite{haidt2004intuitive}, this theory proposes five distinct domains of moral concern rooted in universal evolutionary challenges.
%(e.g., vulnerability of offspring, co-operation and sharing, differences in physical capabilities, disease / parasite control and maintaining kin boundaries).
These lead to five distinct foundations for moral concern: care/harm, fairness/cheating, authority/subversion, sanctity/degradation, and loyalty/betrayal. 
A key application case of this model is explaining political differences -- liberals and conservatives endorse these moral foundations differently \cite{haidt2007morality, graham2009liberals}.

\begin{table*}[ht!]
\scriptsize
\centering
\begin{tabular}%{p{12mm}|p{79mm}|p{11mm}|p{11mm}|p{11mm}|p{11mm}|p{11mm}}
{c>{\raggedright\arraybackslash}m{70mm}ccccc}
% \begin{tabular}{|c|L|c|c|}
 \hline
 \textbf{Stance} & \multicolumn{1}{c}{\textbf{Tweet}} & \textbf{Care} & \textbf{Fairness} & \textbf{Loyalty} & \textbf{Authority} & \textbf{Sanctity} \\
 \hline
 FAVOR & @HillaryClinton the @DalaiLama speaks of women in leadership roles bringing about a more compassionate world. \#potus \#SemST & + & + & + & + & + \\
 \hline
 FAVOR & Just met an awesome supporter on the CX bus! He said "Hillary is one strong woman and we need that for our country." \#FellowsNV \#SemST & + & -- & + & + & + \\
 \hline
 AGAINST & I wish \#OliviaPope could run @HillaryClinton 's campaign... \#Scandal \#livisreal \#SemST & + & -- & + & -- & -- \\
 \hline
 AGAINST & Why did you lie about the \#Benghazi subpoena? @HillaryClinton No wonder no one trusts you. \#SemST & -- & -- & -- & + & + \\
 \hline             
 NONE & @larryelder The more Republicans talk about social policy the better....for \#SemST & -- & -- & + & + & + \\
 \hline
\end{tabular}
\caption{Sample tweets for the stance target \textit{Hillary Clinton} and their expressed bias towards each moral foundation.}
\label{tab:sample_tweets}
\end{table*}

%\textcolor{red}{Prasanta: This para highlights (a) \textbf{how} we incorporate moral foundations into the stance detection models, and (b) a preview of our key findings} 
Given the important role that moral foundations play in shaping social behavior, we posit that they also enable the formation of human opinions and stances.
TABLE~\ref{tab:sample_tweets} presents a few tweets from the SemEval 2016 Task 6A dataset~\cite{mohammad2016semeval}, which is widely used for stance detection tasks, demonstrating that it is common for tweets across stance classes to express a positively- or negatively-valenced bias towards above-mentioned moral foundations. Hence, we hypothesize that the systematic extraction and incorporation of these moral foundations should enhance the detection of both, message- and user-level stances from social media-based content. In testing this hypothesis, we seek to augment online stance detection tasks on Twitter datasets by incorporating moral foundation features into message- and user-level stance detection models. Our findings reveal that the addition of moral foundation features boosted F1 score of stance detection models by up to 23.7 points, depending on the choice of model and dataset. %For models with advanced features (e.g., Sentence-BERT), the improvements from adding moral foundations were predictably more modest, with an average improvement of around 1.5-1.6\% depending on the choice of model and dataset. 
Beyond improving the predictive performance on stance detection, the insights generated from our association-based analyses highlight the prevalence and nature of moralized discourse surrounding key targets (e.g., wearing masks) across datasets. 
%Notably, our association-based analyses highlight the prevalence of certain moral foundation attributes (e.g., sentiment and bias) in tweets on specific topics (e.g., wearing mask). 
%The insights derived from our analyses contribute to characterizing the moralized nature of social media discourse surrounding these topics.  
%We extract moral foundation from social media data at both text and user level, to measure moral values for each foundation, and use the moral value embedding to perform stance detection.
%\textcolor{red}{Proposed rephrasing for this paragraph:}
%\textcolor{blue}{
%Due to the role that moral foundations play in such processes, we believe that moral foundations can also help to reveal human opinions and stances. In this paper, we test this hypothesis by leveraging moral foundations to infer human opinions and stances from text. Specifically, we compute moral foundation scores from social media data at both the text and user level, and include these scores as additional features for training text-based stance-detection models. Our results show that (DESCRIBE FINDINGS HERE), and demonstrate the importance of including moral foundation dimensions in stance detection tasks.}

%\textcolor{red}{Prasanta: This section starts with summarizing the importance and implications of this work. Followed by a introduction to the upcoming section(s).}
Through this study, we offer four key contributions. Firstly, and as previously mentioned, current stance detection models are limited by their reliance on conventional textual and contextual features. We performed a comprehensive study to assess the predictive utility of moral foundation within the context of stance detection task on social media data.
%While some recent studies have explored the relevance of moral foundations through correlational analysis \cite{rezapour2021incorporating} or qualitative evaluations \cite{roy2021analysis} in the domain of online stance, we believe that this is among the first attempts to assess the predictive utility of moral foundations within the context of stance detection task on social media.
Secondly, we highlight important heterogeneities in the predictive performance of moral foundation-based models across tasks (i.e., message-level vs. user-level stance detection), datasets, stance targets, and classifiers. 
Thirdly, in addition to predictive performance, this study highlights interesting associations between moral foundations and the stances towards specific targets. 
%This helps to emphasize the prevalence of moralized content in social media discourse that is framed in favour or against specific target topics.
Lastly, we show that the incorporation of moral foundations improves F1 score on stance detection and prediction tasks on average by 1.06 points for traditional machine learning (LM) models, 5.91 points for Fine-tuned Language Model (FLM), and 15.82 points for the more recent Large Language Models (LLM). %This illustrates the importance of modeling psychological attributes of users in such tasks, and 
This highlights that the addition of such psychological attributes might be particularly fruitful for LLM-based stance detection models. Further research can draw on these insights to explore the design of psychologically-rooted LLMs for related tasks.

%In the next section, we briefly review the main stance detection techniques that leverage various features and models, and the various methods of extracting moral foundations, focusing primarily on the MFD and the extended MFD (eMFD). Then, we introduce our modeling strategies andexperiments based on text- and user-level dataset. We conclude with key findings and a discussion on the future direction of this work.

\section{Related Work}

\subsection{Stance Detection}
When expressed in text, the stance of the message can typically be labeled into any of the constituent classes, e.g., as ``Support'', ``Against'' and ``Neutral''. Stance detection aims to automatically determine the position of a message or its author towards a given proposition or target \cite{mohammad2016semeval}. 
In target-specific stance detection tasks, the models are trained for a particular target \cite{alturayeif2023systematic}. 
%However, such models can also be trained simultaneously towards multiple targets, as in multi-target or cross-target stance detection \cite{sobhani2017dataset, wei2018multi}. 
However, more recent work has investigated the problem of zero-shot stance detection which is target-agnostic and aims to detect the stance towards new or unseen targets \cite{allaway2020zero}.
As a language modeling task, stance detection has been widely studied in contexts spanning politics \cite{taule2018overview,zhang2023wearing}, 
%rumor and fake news \cite{thorne2017fake, umer2020fake}, 
climate change \cite{upadhyaya2023multi}, and the COVID-19 pandemic \cite{glandt2021stance}.

%\textcolor{red}{Prasanta: CITE 1-2 health related studies. stance towards vaccines/masks etc.}

%The most common definition of a stance is that it represents an external or expressed manifestation of our opinions and beliefs \cite{du2007stance,biber1988adverbial}. When expressed in text, the stance of the message can typically be labeled into any of the constituent classes, e.g., as ``Support'', ``Against'' and ``Neutral''. Subsequently, stance detection models can be trained for a particular target, a modeling technique known as target-specific detection \cite{alturayeif2023systematic}. It can also be trained simultaneously towards multiple targets, using multi-target or cross-target stance detection \cite{sobhani2017dataset, wei2018multi}. More recent work has investigated the problem of zero-shot stance detection which is target-agnostic and aims to detect the stance towards new or unseen targets \cite{allaway2021adversarial,allaway2020zero}. 
%A less popular, but useful, stream of work focuses on stance discovery using unsupervised approaches \cite{darwish2020unsupervised,kobbe2020unsupervised,samih2021few}, and some have been used successfully in user stance detection problems \cite{darwish2020unsupervised,samih2021few}. 
%\textcolor{red}{Prasanta: Add some discussion about target-agnostic stance detection methods, also a bit about papers on stance discovery.}

\subsection{Moral Foundation Dictionary (MFD)}
Moral Foundation Theory (MFT) \cite{haidt2004intuitive, haidt2007new} posits five moral foundations that are prevalent across cultures and nations: 
Care/Harm, Fairness/Cheating, Loyalty/Betrayal, Authority/Subversion, and Sanctity/Degradation. 
To measure these five dimensions from text, \citet{graham2009liberals} developed the first MFD using a two-phase approach. The first phase involved generating associations, synonyms and antonyms for the five moral foundations through thesauruses and conversations with peers. Next, words that were too distantly related to the foundations were removed. The result of this exercise was a dictionary of 295 words related to five moral foundations, that are also used in computational linguistic tools such as the Linguistic Inquiry and Word Count program (LIWC) \cite{graham2009liberals}.
Based on the foundation of MFD, \citet{rezapour2019enhancing} expanded the morality lexicons and it was used to understand the usefulness of lexical auxiliary tools. The expanded dictionary with 4,636 words was studied in the task of measuring social effects. 

\subsection{Extended MFD (eMFD)}
Although the MFD offers an automatic and dictionary-based method to extract moral foundation cues from text, it has a few limitations \cite{ garten2018dictionaries, weber2021extracting}. For instance, MFD was created by a small group of \textit{experts} which might limit its generalizability in a broader and \textit{non-expert} population. Moreover, the MFD adopts a ``winner takes all'' strategy, which assigns a word to a single moral foundation, precluding the possibility of a word being related to multiple moral foundations at once. To address these concerns, \citet{hopp2021extended} proposed the extended MFD (eMFD) to help capture large-scale and intuitive moral judgments from text. Instead of relying on careful selection by a small group of experts, the eMFD lexicon was generated through a wider crowd-sourced task aimed at capturing a more comprehensive list of morally relevant content cues. Moreover, instead of a single moral dimension, the eMFD assigns each word a vector of scores, which reflects the probability of that word belonging to each moral foundation.

%\textcolor{red}{I commented out the content of VADER as it's a sentiment model. We should discuss closely related papers that leverage moral foundations for stance models, such as \cite{rezapour2021incorporating} and \cite{upadhyaya-etal-2023-toxicity}. In introduction, \cite{rezapour2021incorporating} is mentioned as a few recent exceptions, but never discussed with any detail. That creates points of being attacked by reviewers easily.}.
%Unlike MFD which assigns words to binary categories of virtue and vice,  \citet{hopp2021extended} utilized a continuous approach by capturing the valence of context using the Valence Aware Dictionary and sEntiment Reasoner (VADER) \cite{hutto2014vader}. The resulting valence sentiment scores range from -1 to 1, with -1 being the highest vice and 1 being the highest virtue.

\subsection{FrameAxis}
Moral foundation dictionary is only effective when the corpus of interest contains words from the dictionary. However, moral information in text can be expressed using diverse linguistic cues and styles, which might include only few or none of the words present in the MFD. In such contexts, it is challenging to infer the underlying moral foundations effectively using just a dictionary-based approach. 
%A recent method on \textit{framing analysis} by \citet{kwak2021frameaxis} offers a possible resolution to this problem. Framing can help with highlighting a particular aspect of an issue in order to make it salient. In their paper, 
\citet{kwak2021frameaxis} proposed FrameAxis, a method for discovering the presence of framing and its associated biases from documents, by identifying the most relevant semantic facets. Using FrameAxis, any text can be projected on to a high dimensional space using word embeddings to extract moral information, even when none of the words are present in the MFD.
%\textcolor{red}{Comment: It's a bit unclear on the relationship between the ``Recent work on framing analyses'' and the latter mentioned paper by Kwak et al.~\cite{kwak2021frameaxis}. Is it talking about the same paper or different ones? If not same, recent work on framing analyses needs some backup references. If same, the above text need be rephrased to avoid such confusion.}

\section{Datasets}

We use three datasets, which we refer to as the SemEval, Connected Behaviour (CB), and P-Stance datasets. The SemEval dataset was constructed for SemEval 2016 Task 6 \cite{mohammad2017stance}. The dataset contains 4,870 English tweets across six common targets, ``Atheism'' (AT), ``Climate Change is a Real Concern'' (CC), ``Feminist Movement'' (FM), ``Hillary Clinton'' (HC), ``Legalization of Abortion'' (LA), and ``Donald Trump'' (DT). 

The P-Stance \cite{li2021p} dataset is another widely studied stance detection dataset in the political domain. It contains 21,574 labeled tweets across 3 targets, namely ``Donald Trump'' (DT), ``Joe Biden'' (JB), and ``Bernie Sanders'' (BS).
%\citet{mohammad2016semeval} used an SVM classifier with word and character n-grams as features to build a baseline model. Since then, many studies have used this dataset across models and contexts \cite{zhou2017connecting, du2017stance}. 
%For example,  proposed a gated structure on the basis on the biGRU-CNN model by integrating tweet and target embedding, and \citet{du2017stance} developed TAN, a model that combines RNN with long-short term memory (LSTM) and target-specific attention extractor.

The CB dataset is a large Twitter dataset for \emph{user-level} stance detection comprising over 100 million tweets from \cite{zhang2023wearing}. %\citet{zhang2023wearing} attempted to understand how indviduals' future action (e.g., stance on unseen targets) could be predicted based on their past behaviors. 
This dataset was used for user-level stance detection towards three targets, namely ``Donald Trump'' (DT), ``Wearing Mask'' (WM), and ``Racial Equality'' (RE). 
% Tweets were collected separately for each target using carefully curated keywords.

\section{Our Method}

\begin{figure*}[t!]
  \centering
  \begin{subfigure}[b]{0.48\textwidth}
    \centering
    \includegraphics[width=\textwidth]{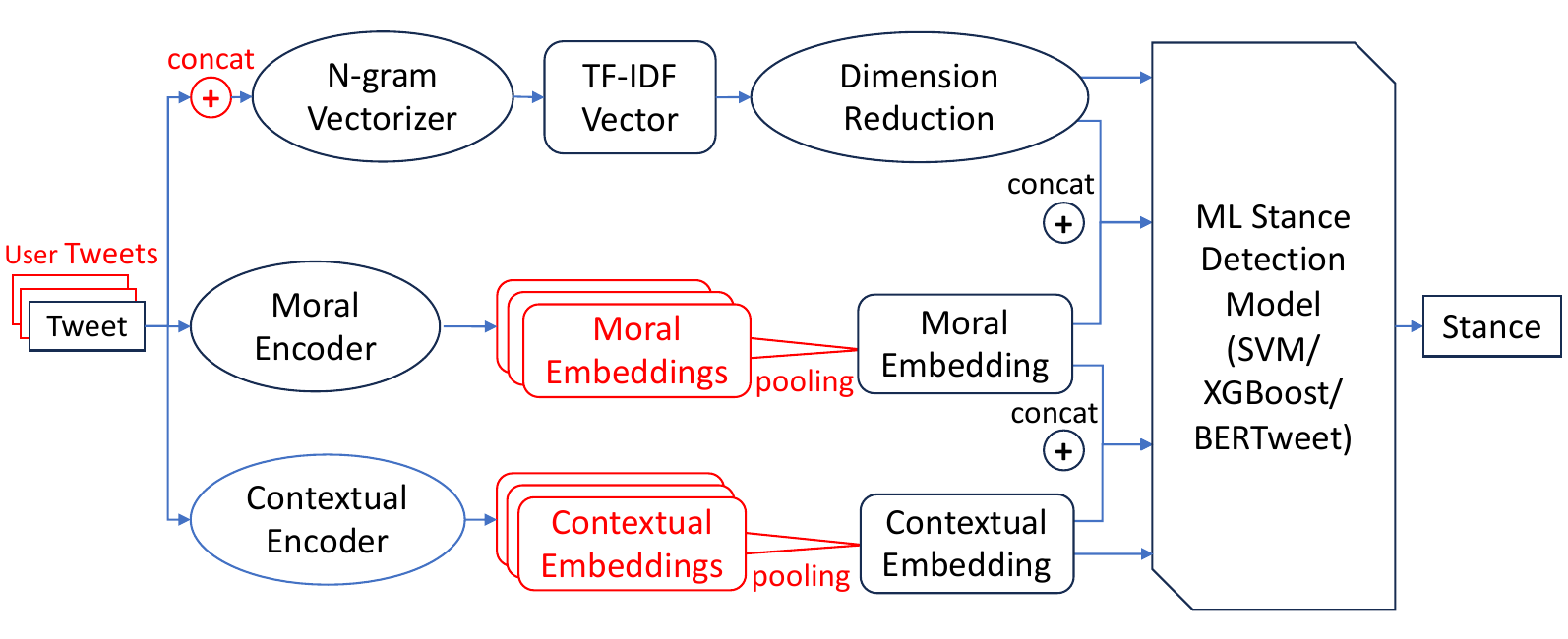}
    \caption{With traditional stance detection model}
    \label{fig_flowchart_ml}
  \end{subfigure}
  \hfill
  \begin{subfigure}[b]{0.48\textwidth}
    \centering
    \includegraphics[width=\textwidth]{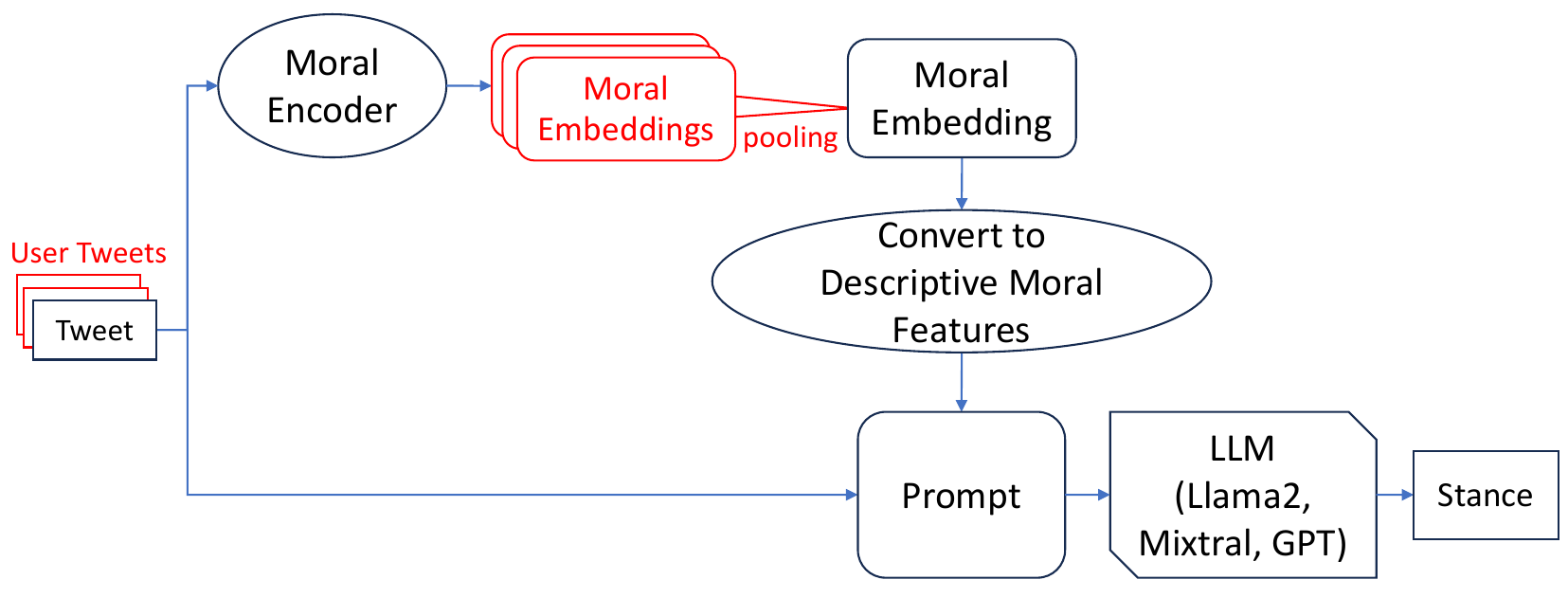}
    \caption{With LLM-based stance detection model}
    \label{fig_flowchart_llm}
  \end{subfigure}
  \caption{Our method for enhancing stance classification with quantified moral foundations based on traditional ML models and LLMs. Components in red were only used for user-level stance detection, where tweets posted by same user were concatenated before passing through TF-IDF vectorizer, and pooling was applied for moral and contextual embeddings.}
  %\textcolor{red}{the figures only considers post-level stance? How about the user level?}\textcolor{blue}{Tommy: added a section "User-level Feature Aggregation" to explain about user-level feature pooling. Also added another chart for user-level architecture, let's discuss on monday whether we should use the user-level chart}}
  \label{fig:flowchart_all}
\end{figure*}

% \begin{figure*}[t!]
%   \centering
%   \begin{subfigure}[b]{0.48\textwidth}
%     \centering
%     \includegraphics[width=\textwidth]{figures/user_ml_flowchart.pdf}
%     \caption{With traditional ML-based user-level stance detection model}
%     \label{fig_flowchart_ml}
%   \end{subfigure}
%   \hfill
%   \begin{subfigure}[b]{0.48\textwidth}
%     \centering
%     \includegraphics[width=\textwidth]{figures/user_llm_flowchart.pdf}
%     \caption{With LLM-based user-level stance detection model}
%     \label{fig_flowchart_llm_user}
%   \end{subfigure}
%   \caption{Our method for enhancing user-level stance classification with quantified moral foundations based on traditional machine learning and LLM. A tweet is first concatenated as a user document before generating TF-IDF (with reduced dimensions using PCA and UMAP). Each tweet is passed through seperate channels in parallel to generate moral and contextual embeddings before pooling layer. We combine pooled moral embedding with TF-IDF, pooled contextual embeddings, or LLM in different ways for user-level stance detection.}
%   \label{fig:entire}
% \end{figure*}

In this section, we discuss how morality was encoded and injected into both traditional ML models and LLMs. The gain in performance of the stance detection task with the addition of moral features was studied using the design shown in Fig.~\ref{fig:flowchart_all}. 

%\textcolor{red}{Please add a paragraph generally describing the approach or any motivational points of the proposed experimental approach/framework, as the preface of the entire section, to bridge the gap.}

\subsection{Feature Encoding}
To incorporate moral features with stance detection models based on distinct learning methods, we used TF-IDF, contextual, and moral embedding to represent features. As shown in Fig.~\ref{fig:flowchart_all}, tweets were passed through separate channels to generate these three embeddings with reduced dimensions using PCA and UMAP. These embeddings were then combined (converted to descriptive moral features for LLM case) in different ways for stance detection.
%To ensure our findings are consistent over different types of encoding methodology, we consider different types of features. %Traditional language models often rely on a bag-of-words approach using words sequence, known as n-grams, as features. 

\textbf{TF-IDF.} The term frequency-inverse document frequency (TF-IDF) vectorizer represents n-gram features by using both character- or word-level TF-IDF values. %uses overlapping characters and word n-grams as a vector that reflects the importance of specific characters and words in the corpus. %This has been applied in many studies spanning document retrieval, text classification and stance detection \cite{shannon1948mathematical, furnkranz1998study, popovic2015chrf, ghanem2018stance}. 
%In this study we consider n-gram features by applying both character and word TF-IDF as the first type of baseline features. 

\textbf{Contextual Embedding.} Sentence-BERT~\cite{reimers2019sentence} is a variant of Transformer-based models~\cite{vaswani2017attention} that uses siamese and triplet network structure in the training stage and has been shown to outperform BERT and RoBERTa for sentence embedding tasks \cite{liu2019roberta}. Here we encoded tweets with the SBERT all-mpnet-base-v2 version to generate contextual embeddings.
%following the comparative study on stance detection done by~\citet{ghosh2019stance}.

\textbf{Moral Embedding.} Two separate techniques were used to generate moral embeddings. The first was the \textbf{eMFD embedding} \cite{hopp2021extended}. 
%To produce this embedding, each word in the input text was matched against the words in the eMFD dictionary to retrieve five moral foundation \emph{probabilities} and five foundation \emph{sentiment} scores, and then, these probabilities and sentiment scores were averaged to obtain a 10-dimensional embedding. 
Using this method, \emph{probability} and \emph{sentiment} scores for a text along a moral foundation dimension were computed based on the frequency of occurrence of eMFD keywords in the text.
This produced a 10-dimensional vector for each text (5 probabilities and 5 sentiment scores).
% \footnote{\url{https://github.com/medianeuroscience/emfdscore}}
%The second technique is called \textbf{FrameAxis embedding}. By considering the microframe vector for pair of antonyms (e.g. care-harm) as \(v_f=v_{w+}-v_{w-}\), where \(v_{w+}\) and \(v_{w-}\) are the word vector of pole words, \citet{kwak2021frameaxis} defined the contribution of a word \(w\) on a microframe \(f\) as the cosine similarity between their vectors. Microframe bias of a corpus on a microframe \(f\) is hence the weighted average of each word's contribution to \(f\) for all words in the corpus. Microframe intensity captures the strength of a microframe in the document and is measured by the second moment of word contributions on microframe \(f\) for all words in the document.\textcolor{red}{Reviewer without knowing FrameAxis will be completely lost in this paragraph -- description unclear.}
% \footnote{\url{https://github.com/negar-mokhberian/Moral\_Foundation\_FrameAxis}}
The second was the Moral Foundation \textbf{FrameAxis embedding} \cite{mokhberian2020moral}, which combined the eMFD with the FrameAxis method described in \cite{kwak2021frameaxis}.
In this method, the \emph{bias} and \emph{intensity} of a text along a moral foundation dimension are functions of the cosine similarity of word embeddings from the text to word embeddings of eMFD keywords for that moral dimension.
This produced a 10-dimensional vector for each text (5 bias and 5 intensity scores).

\textbf{User-level Representation.} The CB dataset targets user-level stance tasks. 
%\textcolor{blue}{We haven't introduced the datasets yet, so probably shouldn't mentioned the CB dataset here. Maybe write something like: "For user-level stance detection tasks, feature encodings were aggregated...."}\textcolor{teal}{Tommy: moved Dataset section up so we can refer to them thereafter}. 
Feature encodings were aggregated to form user-level representations. As TF-IDF is a statistical method relying on word and document frequency, tweets posted by the same user were first concatenated to obtain a document for each of the targets. These user documents were then used to calculate TF-IDF embeddings. As other feature encodings have a fixed dimension, we applied mean-pooling to obtain user-level embeddings from tweet embeddings.
%\textcolor{red}{There is design choice issues: why other feature encodings didn't follow the same way as TF-IDF by first concatenating the tweets and applying say SBERT?}.

\subsection{Stance Detection Models}
We compared three broad classes of models for the stance detection task.

\subsubsection*{1) Traditional Machine Learning Models}
% In this paper, we train SVM~\cite{noble2006support} and XGBoost~\cite{chen2016xgboost} classifiers using both n-gram and SBERT embeddings as basic features and augment these models by incorporating morality features through moral embeddings. Such models have been widely used in many stance detection studies~\cite{mohammad2016semeval, augenstein2016stance, kuccuk2018stance, chen2016xgboost}.

In this study, we trained SVM and XGBoost classifiers using both n-gram and SBERT embeddings as basic features, and augmented these models by incorporating morality features through moral embeddings. Such models have been widely used in many stance detection studies~\cite{mohammad2016semeval, chen2016xgboost}.

\subsubsection*{2) Fine-tuned Language Models}
%\textcolor{red}{Should we also have a latest PLM-based model such as Roberta which is believed stronger than SVM/Boost model?}

%\textcolor{brown}{\emph{I have changed Pre-Trained Language Models to Fine-Tuned Language Models instead, as BERT models can only do stance classification after being fine-tuned on it. Please verify if this is an accurate description of what was done.}}

Fine-tuned language models (FLMs) such as BERT-based models have been identified as state-of-the-art (SoTA) in stance detection tasks~\cite{barbieri-etal-2020-tweeteval, li2021p}. In this framework, pre-trained BERT-based models are fine-tuned on a stance classification dataset.

We fine-tuned a pre-trained language model -- BERTweet model~\cite{nguyen-etal-2020-bertweet} on the SemEval, Connected Behavior, and P-Stance datasets. BERTweet is the first public large-scale pre-trained language model for English tweets, having the same architecture as BERT-base~\cite{devlin-etal-2019-bert}. Previous experiments show that BERTweet outperforms strong pre-trained language models, being SoTA models on tweet-based stance tasks~\cite{delucia-etal-2022-bernice}. 
% RoBERTa, XLM-R-base~\cite{conneau-etal-2020-unsupervised}, and other BERT-based models specified for Twitter

% , which follows the RoBERTa pre-training procedure~\cite{zhuang-etal-2021-robustly}

% such as: XLM-T~\cite{barbieri-etal-2022-xlm}, TwHIN-BERT~\cite{zhang2023twhin}, and Bernice~\cite{delucia-etal-2022-bernice}
% We also fine-tuned a pre-trained LongFormer model \cite{Beltagy2020Longformer} on the CB dataset.
% As the CB dataset contained multiple tweets from each user, LongFormer was deemed more suitable given the overall sequence length.

%\textcolor{blue}{Same comment as earlier - shouldn't mention CB dataset without introducing it.}\textcolor{teal}{Tommy: moved Dataset Section to fix this.}

\subsubsection*{3) Large Language Models}
%\textcolor{brown}{\emph{I have re-phrased this section as well}}
We carried out zero-shot and few-shot stance classification using LLMs. 
%Unlike the classes of models, in zero-shot and few-shot stance classification, the LLM is not fine-tuned on the classification task. Instead, the LLM is prompted with tweets from the test dataset and asked to classify the stance based on these inputs. In zero-shot classification, no additional tweets or labels are supplied. In few-shot classification, a few tweets and labels from the training set are supplied, to provide examples of each of the stance classes.
We made use of three prominent LLMs: Llama2-70-chat\footnote{\url{https://replicate.com/meta/Llama2-70b-chat}}, Mixtral-8x7B\footnote{\url{https://replicate.com/mistralai/mixtral-8x7b-instruct-v0.1}}, and GPT-3.5-turbo\footnote{\url{https://platform.openai.com/docs/models/gpt-3-5-turbo}}.
Llama2-70-chat is based on the Llama 2 family of LLMs \cite{touvron2023llama}, and is fine-tuned for dialogue. 
Mixtral-8x7B is an LLM with a novel sparse mixture of experts (SMoE) architecture \cite{sanseviero2023moe}. GPT-3 \cite{brown2020language} is an autoregressive language model with 175 billion parameters, 10x more than any previous non-sparse language model%, and test its performance in the few-shot setting.
We chose these LLMs as they perform well on many NLP benchmarks while remaining cost-effective.
Their pre-training data also includes social media posts, making them well-suited to process our tweet dataset.
The LLMs were accessed via the Replicate API. Our prompting methods are presented in Section~\ref{section:llm}.

\section{Prompting LLMs with Moral Features}\label{section:llm}
%\textcolor{brown}{\emph{Re-worked this section (5.1 and 5.2) based on what I understand of your description. Please check for accuracy.}}
%In this section, we describe our procedure for generating descriptive moral features and including these features in zero-shot and few-shot stance detection.

\subsection{Generating Descriptive Moral Features}
As moral embeddings are represented as vectors of numbers, it is challenging for LLMs to interpret them. To address this challenge, we employed a two-step methodology: we first clustered these scores and then converted the scores into textual expressions to be included in LLM prompts. Our method is described in Fig.~\ref{fig_flowchart_llm}.

\textbf{Discretization by Clustering}: For each dimension of the eMFD or FrameAxis embeddings, we applied K-Means clustering with $K = 2$ on all the scores for that dimension to determine the threshold between a \emph{high} and a \emph{low} score for that dimension.
This allowed us to categorize each numerical score as either ``High'' or ``Low''.
%Initially, we utilized KMeans clustering for each Morality Feature. Following this, the centroids derived from the clusters were employed to establish the thresholds for each Morality Feature across the datasets. These thresholds were categorized as either low or high.

\textbf{Converting to Text}: Based on whether each tweet was ``High'' or ``Low'' on each moral dimension, we generated textual moral descriptions for that tweet. These descriptions were then included in the stance detection prompts.
% Refer to Appendices \ref{sec:appendix_emfd} and \ref{sec:appendix_frameaxis} for examples of these prompts.

% \textcolor{red}{basically, it's unclear what k-means is applied upon. It's a clustering algorithm, how are the tweets input? Figure shows there is only one tweet input. If it's only one tweet, what is about to be clustered? It's seriously lack of detail, making people quesiton the purpose of it.}

% \subsubsection{FrameAxis features}

% \subsubsection{eMFD features}

\subsection{Prompting for Stance Detection}
%To assess the effectiveness of incorporating morality features from Large Language Models (LLMs) in stance detection, we implemented three distinct prompting schemes. These schemes are designed hierarchically, with each subsequent scheme building upon the information integrated in the preceding one.
For stance detection with LLMs, we implemented three distinct prompting schemes. These schemes were designed hierarchically, with each subsequent scheme built upon the information integrated in the preceding one.
%The following list details each of the prompting schemes we used to classify stance for each of the dataset.

\textbf{Task + Context.} The prompt included the task description and necessary contextual information, the stance target, and a clear definition of stance labels. This is consistent with previous studies leveraging ChatGPT for stance labeling~\cite{aiyappa-etal-2023-trust}.

\textbf{Task + Context + FrameAxis.} We further augmented the Task + Context prompt by including the moral descriptions generated from the FrameAxis embeddings. 

\textbf{Task + Context + eMFD.} Alternatively, we augmented the Task + Context prompt by including the moral descriptions generated from the eMFD embeddings.

For the latter two schemes, we included the moral descriptions along with explanations of each moral dimension. Our hypothesis was that an LLM could make better stance predictions by considering these moral features.

Our prompting strategies are inspired by the template samples available in HuggingFace's resources for LLaMa2\footnote{\url{https://huggingface.co/blog/llama2}} and Mixtral\footnote{\url{https://huggingface.co/mistralai/Mixtral-8x7B-Instruct-v0.1}}.
We implemented zero-shot, 1-shot, and 5-shot classification scenarios for the 3 schemes mentioned above. 
% Detailed specifications of the prompts used for each dataset can be found in the Appendix~\ref{appendix:prompt_detail}.

\section{Experiments and Results}
%In the following section, we present our experimental setups, comprehensive examinations of our experiments and the results obtained from these investigations.

\subsection{Experimental Setup and Data Sampling}

%\textcolor{brown}{\emph{Moved the hyper-parameter section here. It should be part of the methods, not experimental results}}
We tuned the hyperparameters for SVM and XGBoost following~\citet{gera2022comparative}. We performed grid search on SVM with two kernels. The parameters \{1e-3, 1e-4\} for gamma, and \{1, 10, 100, 1000\} for \textit{C} were searched with the Radial Basis Function (RBF) kernel, and \{1, 10, 100, 1000\} for \textit{C} with the linear kernel. Similarly, we performed hyperparameter tuning for XGBoost using grid search in \{100, 500\} for \textit{n\_estimators}, \{5, 10, 15, 20\} for \textit{max\_depth} and \{0.01, 0.1\} for \textit{learning\_rate}. We applied a stratified 2-fold cross validation in the search process. 

%The number of records for each stance class across the two datasets is shown in Tables ~\ref{tab:semeval_class_imbalance} and ~\ref{tab:cb_class_imbalance}. In Table~\ref{tab:semeval_class_imbalance}, 
The stance classes in SemEval 2016 dataset are heavily imbalanced. 
%For example, in the extreme case, 212 tweets from the target ``Climate Change is a Real Concern'' have a ``Favor'' stance, while only 15 have an ``Against'' stance. 
%This might result in a naive classifier where the features might not be strong enough to generate a good performance. Furthermore, the use of F1 score as a performance metric is problematic under the influence of high class imbalance. To alleviate this problem, 
We applied oversampling to balance training examples from different classes. 
In the Connected Behavior dataset, we sampled 5 tweets from each of the top 500 active users, based on the number of posted tweets for each target and stance class, 
%\textcolor{red}{what makes user a top user? according to \# of posts? Clarify please.} 
to avoid having an extremely large and sparse matrix due to the size of the vocabulary for TF-IDF. We did not observe a strong class imbalance for the P-Stance dataset, and hence,  sampled 1,000 instances from the training data of each target for the same purpose. %when applying TF-IDF vectorizor.
%\textcolor{red}{Don't understand this sentence: what exactly are sampled from training data -- tweets? 1000 for each class or all classes?}.   %This successfully reduced the problem of naive classification scenarios, and improved model performance for most cases. 
% Undersampling could not be applied in this case due to the limited size of the SemEval dataset.

We followed \citet{nguyen-etal-2020-bertweet} to fine-tune BERTweet for each dataset and task over 30 epochs. We used AdamW with a learning rate of 1.e-5 and a batch size of 32, and  assessed performance after each epoch with early stopping applied if no improvement occurred over 5 epochs. The best checkpoint was selected for test set evaluation.
For prompting-based experiments, we set each of the models to only provide the most probable output (e.g., by setting \texttt{temperature=0)} and a maximum length of 5 tokens \cite{cruickshank2023use, 10.1145/3628797.3628837}. %Each experiment was carried out five times, and the average F1 score is reported as the final score for prompting-based LLMs experiments.
Each experiment was carried out ten times with random seeds, and the average F1 score is reported as the final score.

\subsection{Results and Analysis}

\begin{table*}[t!]
\centering
\resizebox{0.80\textwidth}{!}{
\begin{tabular}{l|c|c|c|c} 
\hline
\multirow{2}{*}{\textbf{Features/Schemes}} & \multirow{2}{*}{\textbf{Models}}                                                      & \multicolumn{3}{c}{\textbf{Datasets}}                                                                                    \\ 
\cline{3-5}
                                           &                                                                                       & \textbf{SemEval}                       & \textbf{CB}                            & \textbf{P-Stance}                      \\ 
\hline
n-gram                                     & \multirow{3}{*}{SVM/XGB (+PCA)}                                                       & 0.502/0.465                            & 0.820/0.789                            & 0.671/0.660                            \\
n-gram + eMFD                              &                                                                                       & 0.519/0.481                            & 0.825/0.801                            & 0.679/0.677                            \\
n-gram + FrameAxis                         &                                                                                       & 0.525/0.475                            & 0.822/0.803                            & 0.690/0.688                            \\ 
\hline
n-gram                                     & \multirow{3}{*}{SVM/XGB (+UMAP)}                                                      & 0.420/0.449                            & 0.796/0.763                            & 0.660/0.654                            \\
n-gram + eMFD                              &                                                                                       & 0.442/0.477                            & 0.806/0.773                            & 0.668/0.659                            \\
n-gram + FrameAxis                         &                                                                                       & 0.452/0.478                            & 0.806/0.781                            & 0.674/0.669                            \\ 
\hline
SBERT                                      & \multirow{3}{*}{SVM/XGB}                                                              & 0.622/0.603                            & 0.774/0.774                            & 0.751/0.734                            \\
SBERT + eMFD                               &                                                                                       & 0.625/0.611                            & 0.778/0.780                            & 0.750/0.739                            \\
SBERT + FrameAxis                          &                                                                                       & 0.630/0.614                            & 0.778/0.795                            & 0.749/0.736                            \\ 
\hline
Tweet only                                 & \multirow{3}{*}{FLM (BERTweet)}                                                       & 0.713                                  & 0.655                                  & 0.793                                  \\
Tweet + eMFD                               &                                                                                       & 0.717                                  & 0.663                                  & 0.807                                  \\
Tweet + FrameAxis                          &                                                                                       & 0.753                                  & 0.701                                  & 0.799                                  \\ 
\hline
Task + Context                             & \multirow{3}{*}{\begin{tabular}[c]{@{}c@{}}Llama/Mixtral/GPT\\Zero-shot\end{tabular}} & \multicolumn{1}{l|}{0.541/0.372/0.665} & \multicolumn{1}{l|}{0.435/0.220/0.723} & \multicolumn{1}{l}{0.619/0.226/0.748}  \\
Task + Context + eMFD                      &                                                                                       & \multicolumn{1}{l|}{0.590/0.427/0.666} & \multicolumn{1}{l|}{0.594/0.404/0.743} & \multicolumn{1}{l}{0.694/0.279/0.754}  \\
Task + Context + FrameAxis                 &                                                                                       & \multicolumn{1}{l|}{0.586/0.393/0.668} & \multicolumn{1}{l|}{0.637/0.344/0.740} & \multicolumn{1}{l}{0.806/0.347/0.755}  \\ 
\hdashline
Task + Context                             & \multirow{3}{*}{\begin{tabular}[c]{@{}c@{}}Llama-/Mixtral/GPT\\1-shot\end{tabular}}   & \multicolumn{1}{l|}{0.555/0.392/0.686} & \multicolumn{1}{l|}{0.440/0.248/0.737} & \multicolumn{1}{l}{0.628/0.239/0.779}  \\
Task + Context + eMFD                      &                                                                                       & \multicolumn{1}{l|}{0.590/0.447/0.700} & \multicolumn{1}{l|}{0.615/0.429/0.767} & \multicolumn{1}{l}{0.678/0.292/0.787}  \\
Task + Context + FrameAxis                 &                                                                                       & \multicolumn{1}{l|}{0.595/0.475/0.714} & \multicolumn{1}{l|}{0.677/0.448/0.781} & \multicolumn{1}{l}{0.799/0.393/0.809}  \\ 
\hdashline
Task + Context                             & \multirow{3}{*}{\begin{tabular}[c]{@{}c@{}}Llama/Mixtral/GPT\\5-shot\end{tabular}}    & \multicolumn{1}{l|}{0.590/0.414/0.714} & \multicolumn{1}{l|}{0.467/0.279/0.757} & \multicolumn{1}{l}{0.624/0.281/0.808}  \\
Task + Context + eMFD                      &                                                                                       & \multicolumn{1}{l|}{0.640/0.479/0.744} & \multicolumn{1}{l|}{0.681/0.504/0.792} & \multicolumn{1}{l}{0.683/0.320/0.826}  \\
Task + Context + FrameAxis                 &                                                                                       & \multicolumn{1}{l|}{0.669/0.494/0.759} & \multicolumn{1}{l|}{0.696/0.528/0.810} & \multicolumn{1}{l}{0.808/0.425/0.847}  \\
\hline
\end{tabular}}
\caption{F1-score average across multiple datasets. Llama, Mixtral, and GPT denote Llama2-70b-chat, Mixtral-7x8B, and GPT-3.5-turbo, respectively. CB denoted Connected Behavior dataset.}
\label{tab:exp_results}
\end{table*}

It is important to highlight that the goal of this study is not to produce a state-of-the-art stance detection model that can outperform existing stance detection benchmarks. Instead, we aim to study the effectiveness of encoding moral foundations in improving performance on stance detection tasks across a variety of models and datasets. 
The experiment results are reported in TABLE~\ref{tab:exp_results}. 
%Similar to \citet{gera2022comparative}, we observe the lowest F1 score for the target ``Climate Change is a Real Concern'' among the five targets in SemEval 2016 Task 6A.

\subsubsection{N-gram Baseline}
Using n-gram features as a baseline, the addition of moral features led to an improvement in the F1 score on all datasets, with UMAP as a dimension reduction method showing greater improvement than PCA. The largest improvement with morality was seen on Semeval dataset where the size was smallest, eMFD and FrameAxis has increased F1 score by an average of 2.07 and 2.35 points respectively, across models and dimension reduction methods.
%On P-Stance dataset, eMFD and FrameAxis have led to 1.01\% and 1.84\% improvement respectively. Lastly, on CB dataset which was used for user-level stance detection, eMFD and FrameAxis have improved F1 score by 0.91\% and 1.1\% respectively.

\subsubsection{SBERT Baseline}
To validate the effectiveness of moral foundation features with contextual embedding, the same experiment was repeated using SBERT embeddings~\cite{reimers2019sentence} as baseline model. Although contextual embeddings have been shown to be a stronger baseline over n-gram, we still observed improvements in stance detection performance on adding moral information. Notably, the addition of eMFD and FrameAxis has increased F1 score by up to 0.80 and 1.10 points respectively for tweet-level stance detection on SemEval dataset, and by up to 0.60 and 2.10 points on user-level stance for CB datset.
%, with a F1 score increase of 0.59\% and 0.94\% with the addition of eMFD and FrameAxis on SemEval dataset, respectively. There was a small improvement observed on P-Stance dataset with 0.18\% and \textcolor{red}{0.01\%} for inclusion of eMFD and FrameAxis, respective. On CB dataset, eMFD and FrameAxis have led to an increased F1 score of 0.47\% and 1.25\% for user-level stance detection.

\subsubsection{Fine-tuned Language Models (FLMs)}
We conducted further experiments with FLMs, as they are recognized for achieving SoTA results in stance detection tasks~\cite{barbieri-etal-2020-tweeteval, li2021p}.
However, this did not hold true across all datasets. 
Specifically, we found that FLM underperformed compared to our baseline model, SBERT, on the CB dataset, while they showed superior performance on the SemEval and P-Stance datasets.
Similar to the results observed with traditional ML methods, we observed a significant improvement in stance detection performance upon integrating moral information into FLM. This integration led to F1 improvements in stance detection accuracy by up to 4.00, 4.60, and 1.40 points for SemEval 2016, CB, and P-Stance datasets, respectively.

\subsubsection{LLMs prompting}
Our analysis revealed insightful trends and outcomes in evaluating the performance of Llama2-70b-chat, Mixtral-8x7B, and GPT-3.5-turbo models across three distinct learning scenarios: zero-shot, one-shot, and five-shot on three datasets. 

\textbf{Zero-shot:} As shown in TABLE~\ref{tab:exp_results}, in the SemEval dataset,  the inclusion of eMFD and FrameAxis led to performance improvements of up to 4.90 and 4.50 points, respectively, for Llama2-70b-chat, and up to 5.50 and 2.10 pointsfor Mixtral-7x8B. In the CB dataset, enhancements were even more pronounced, with eMFD and FrameAxis boosting F1 scores by as much as 15.90 and 20.20 points for the  Llama2-70b-chat model,. For the P-Stance Dataset, improvements using this model reached up to 7.50 points with eMFD and an impressive 18.70 points with FrameAxis, highlighting the substantial benefits of integrating moral foundation features.

\textbf{1-shot:} On the SemEval dataset, eMFD and FrameAxis contributed to F1 score increases of up to 3.50 and 4.00 points for Llama2-70b-chat, 5.50 and 8.30 points for Mixtral-7x8B, and 1.4 and 2.8 points for GPT-3.5-turbo indicating salient enhancements. For the CB dataset, including these features raised F1 scores by up to 17.50 and 23.70 points using the  Llama2-70b-chat model, underscoring the important impact of moral foundation features. In the P-Stance dataset, both moral features boosted F1 scores by 5.00 and 17.10 points using the same model, respectively, showcasing the effectiveness of FrameAxis in capturing moral narratives.

\begin{figure*}[hbt!]
    \centering
    \includegraphics[width=15cm]{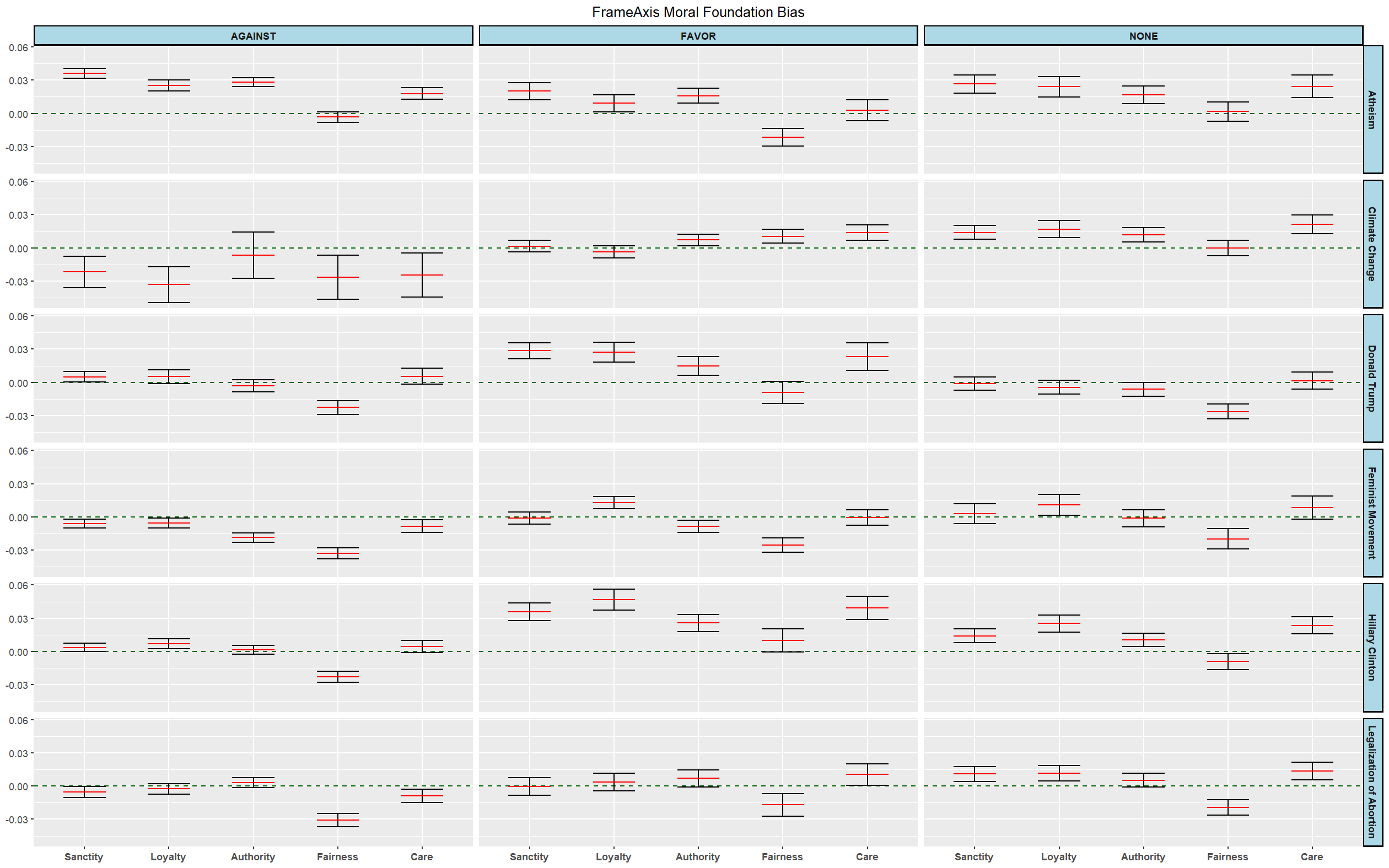}
    \caption{Target- and stance-level heterogeneity in FrameAxis bias of moral foundations from the SemEval 2016 Task 6 dataset.}
    \label{fig:fa_bias_mean_ci}
\end{figure*}

\textbf{5-shot:} In the SemEval dataset, improvements were evident with increases of up to 5.00 and 7.90 points respectively with the addition of eMFD and FrameAxis for Llama2-7b-chat. Similar improvements on the CB dataset were observed using the for Llama2-7b-chat model, with eMFD and FrameAxis leading to increases of up to 21.40 and 22.90 points respectively. In the P-Stance dataset, integrating eMFD and FrameAxis resulted in an increase of 5.90 and 18.40 points respectively, using the Llama2-7b-chat model, illustrating these features' considerable impact.

\section{How do Moral Foundations Affect Stance Conveyance?}

In the previous section, we have highlighted the predictive value of incorporating moral foundations in stance detection tasks. In this section, we take a closer look at the associations between stances towards specific targets, and the expression of specific moral foundations.

In Figure~\ref{fig:fa_bias_mean_ci}, we highlight the prevalence of various FrameAxis bias features for our focal moral foundations across targets and stance classes in the SemEval 2016 dataset. We can observe significant differences across targets, and for different stances within the same target. For example, most people against the target Climate Change is a Real Concern displayed moral violation towards Care, Fairness, Authority and Sanctity foundations, whereas those supporting the target exhibited more righteous moral values.

\begin{figure}[hb!]
    \centering
    \includegraphics[width=\linewidth]{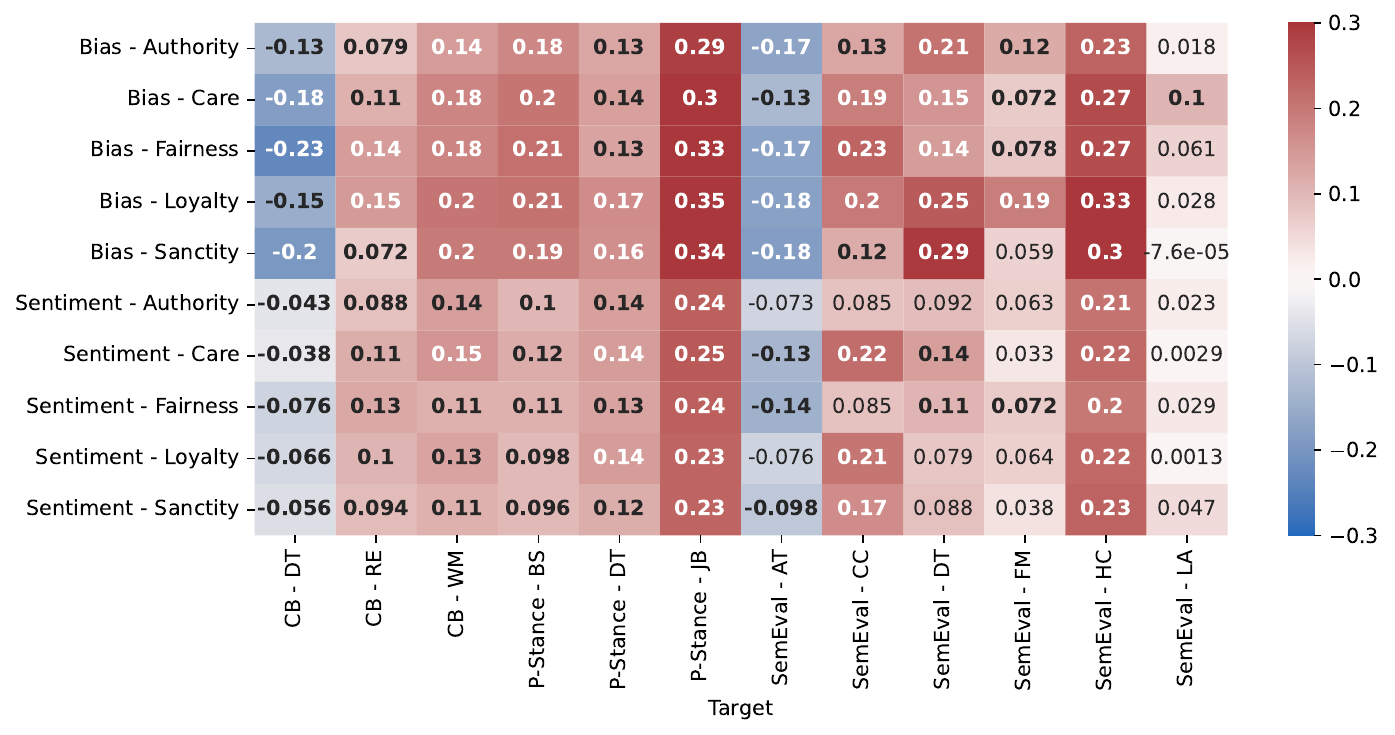}
    \caption{Biserial correlations between moral foundation features and stance. Correlations that are statistically significant at $p<0.05$ significance level are indicated in bold.}
    \label{fig:correlation}
\end{figure}

% \textbf{Add more analysis from the meta-review}\textcolor{red}{Please incorporate the content of Figure 3 putting the figure in the main text. No appendix needed.}

%Specifically, we seek to understand if certain moral positions are over- or under-represented in specific target-stance combinations. 
Moral bias generated from FrameAxis varies significantly across targets, as well as between stance classes within targets. For example, most users against the target \textit{Climate Change is a Real Concern} displayed moral violations towards Care, Fairness, Authority, and Sanctity foundations, whereas those supporting the target exhibited more righteous moral values. 
% More details can be found in Appendix~\ref{sec:appendix_heterogeneity}.

% In Figure~\ref{fig:fa_bias_mean_ci}, we highlight the prevalence of various FrameAxis bias features for our focal moral foundations across targets and stance classes in the SemEval 2016 dataset. We can observe significant differences across targets, as well as between stance classes within targets. For example, most people against the target \textit{Climate Change is a Real Concern} displayed moral violation towards Care, Fairness, Authority and Sanctity foundations, whereas those supporting the target exhibited more righteous moral values.

To further analyse how moral values are correlated with stances, in both SemEval 2016 and the Connected Behavior dataset, we encoded the ``Favor'' or ``Support'' stance as 1, and ``Against'' stance as 0, and subsequently measured the biserial correlation between this binary stance, and the sentiment and bias features for each moral foundation. %\textcolor{red}{Prasanta: We haven't yet defined what sentiment, bias, vice, and virtue measures are for the MFD} 
%\textcolor{orange}{Tommy: added that under related work}. 
The statistical significance based on a two-tailed t-test was calculated for each moral foundation feature and target pair.%, to assess the significance level of correlation.
%To further assess the validity of the reported correlations, we conducted a Z-test to compare the mean moral features between Support and Against classes.

In Fig.~\ref{fig:correlation}, eMFD Sentiment and FrameAxis Bias features exhibited the same direction of correlations for most targets and moral foundations. This suggests that the moral information extracted using eMFD and FrameAxis techniques are consistent. The direction of correlation provides an indication of the moral polarity between individuals supporting a target and those against it. The results suggest that exhibiting a greater emphasis on moral values in texts may be a contributing factor to the observed model enhancements. 
%Notably, the two topics demonstrating the most robust correlations--Hillary Clinton and Climate Change is a Real Concern--likewise exhibit the most significant improvements in SBERT-based models, as shown in Figure~\ref{fig:pct_improvement}.

\section{Conclusion and Future Work}
% \textbf{Rephrase it}
% \textcolor{red}{This section can be significantly shortened.}
Moral foundations play a key role in shaping our social behavior in a variety of contexts. While past studies have explored the prevalence and associations of moral foundations in online discourse, the predictive value of moral foundation representations across a diverse range of targets and datasets have remained understudied. In this paper, we investigate if the inclusion of moral foundations can improve the detection of online stances on social media. While existing stance detection models primarily use textual and interaction-based features, our proposed model highlights the importance of incorporating richer user-level attributes, such as their moral foundations. Our models show improved performance in both message- and user-level stance detection tasks using traditional ML models, FLMs and more recent LLMs. Additionally, we highlight associations between stances and each of the five moral foundations, which can provide useful insights for researchers studying online discourse around societal and political events. 

As these five moral foundations are innate, and universally present among all cultures, the moral encoders used in this paper can also be improved from the incorporation of LLM. For instance, future work could also investigate using LLMs to generate moral foundation features, as well as other related attributes (e.g., personalities and beliefs) to improve performance on stance detection tasks.

% \bigskip
% \noindent Thank you for reading these instructions carefully. We look forward to receiving your electronic files!

% Use \bibliography{yourbibfile} instead or the References section will not appear in your paper
\bibliographystyle{IEEEtranN}
\bibliography{acl2023}

% \newpage
%\onecolumn
% \appendix

\newcommand{\teal}[1]{{\leavevmode\color{teal}#1}}
\definecolor{brown}{rgb}{0.6, 0.4, 0.2}
\newcommand{\brown}[1]{{\leavevmode\color{brown}#1}}
\newcommand{\cyan}[1]{{\leavevmode\color{cyan}#1}}
\newcommand{\purple}[1]{{\leavevmode\color{purple}#1}}

\end{document}